\documentclass[a4paper,fleqn]{cas-dc}

\usepackage[numbers]{natbib}
\def\tsc#1{\csdef{#1}{\textsc{\lowercase{#1}}\xspace}}
\tsc{WGM}
\tsc{QE}
\begin{document}
\let\WriteBookmarks\relax
\def\floatpagepagefraction{1}
\def\textpagefraction{.001}
\let\printorcid\relax % 可去掉页面下方的ORCID(s)

% \shorttitle{<short title of the paper for running head>} 
\shorttitle{Gait Recognition Based on Tiny ML and IMU Sensors}    

% \shortauthors{<short author list for running head>}
\shortauthors{Zhengbao Yang et al.}

\title[mode = title]{Gait Recognition Based on Tiny ML and IMU Sensors}

\author[1]{Jiahang Zhang}%[style=chinese]

\author[1]{Mingtong Chen}%[role=Co-ordinator, suffix=Jr]
%\ead{rishi@sayahna.org}
%\ead[URL]{www.sayahna.org}
%\credit{Data curation, Writing - Original draft preparation}

\author[1]{Zhengbao Yang}
\cormark[1] 
%\fnmark[3]
\ead{zbyang@hk.ust} 
\ead[URL]{https://yanglab.hkust.edu.hk/}

\address[1]{The Hong Kong University of Science and Technology
Hong Kong, SAR 999077, China}
% \address[2]{Sayahna Foundation, Jagathy, Trivandrum 695014, India}
% \address[3]{\TeX{} Users Group, Providence, MA, USA}

\cortext[1]{Corresponding author} 
%\cortext[2]{Principal corresponding author} 

\begin{abstract}
This project presents the development of a gait recognition system using Tiny Machine Learning (Tiny ML) and Inertial Measurement Unit (IMU) sensors. The system leverages the XIAO-nRF52840 Sense microcontroller and the LSM6DS3 IMU sensor to capture motion data, including acceleration and angular velocity, from four distinct activities: walking, stationary, going upstairs, and going downstairs. The data collected is processed through Edge Impulse, an edge AI platform, which enables the training of machine learning models that can be deployed directly onto the microcontroller for real-time activity classification.The data preprocessing step involves extracting relevant features from the raw sensor data using techniques such as sliding windows and data normalization, followed by training a Deep Neural Network (DNN) classifier for activity recognition. The model achieves over 80\% accuracy on a test dataset, demonstrating its ability to classify the four activities effectively. Additionally, the platform enables anomaly detection, further enhancing the robustness of the system. The integration of Tiny ML ensures low-power operation, making it suitable for battery-powered or energy-harvesting devices.

\end{abstract}


% \begin{highlights}
% \item highlight-1
% \item highlight-2
% \item highlight-3
% \end{highlights}

% Keywords

\maketitle

% Main text
\section{Introduction}

Tiny ML (Tiny Machine Learning) is a technology that embeds machine learning models into resource-constrained embedded devices or microcontrollers. It allows machine learning models to perform inference and execution on low-power, low-computation devices without relying on cloud computing. This technology typically involves using frameworks such as TensorFlow Lite, uTensor, and Edge Impulse to compress and optimize pre-trained models to fit the hardware resources of embedded platforms\cite{1,2}.

One significant advantage of Tiny ML is its ability to perform local inference on edge devices, meaning that the device can process data and respond in real time without uploading data to the cloud for processing. This approach reduces latency, saves bandwidth, and enhances data privacy, making it especially suitable for devices that need to operate for long periods with limited power, such as wearables and IoT devices\cite{3,4}.

Deploying Tiny ML models onto microcontrollers can greatly reduce power consumption and enhance device autonomy. Traditional machine learning applications often require substantial computational resources and continuous network connections. However, by integrating Tiny ML into microcontrollers, all computations and inferences can be performed locally, eliminating the need for external server dependency. This can significantly extend battery life, making it ideal for battery-powered or energy-harvesting devices such as smartwatches and health monitoring systems.

Furthermore, running Tiny ML on microcontrollers offers extremely low latency, as data no longer needs to be uploaded to the cloud for processing, avoiding network delays. This allows for faster and more responsive systems, making it suitable for applications that require quick reactions, such as motion monitoring and environmental sensing.

The goal of this project is to develop a gait recognition system based on Tiny Machine Learning and IMU sensors. By using the XIAO-nRF52840 Sense microcontroller and the LSM6DS3 IMU sensor, combined with the Edge Impulse platform for data collection, processing, and model training, the system aims to classify four common activities (Walking, Stationary, Going Upstairs, Going Downstairs) in real time\cite{5}.

In this project, the IMU sensor I used is the LSM6DS3. It is a six-axis inertial sensor developed by STM icroelectronics, integrating a three-axis accelerometer and a three-axis gyroscope. It provides high-precision measurements of acceleration and angular velocity, making it widely used in motion detection, posture recognition, and positioning/navigation applications. The LSM6DS3 supports programmable output data rates and low-power operating modes, making it ideal for real-time data collection and processing in resource-constrained devices.

This sensor has multiple operating modes, allowing users to select the appropriate balance between power consumption and performance based on specific needs. In low-power mode, it extends battery life, while in high-sampling-rate mode, it can provide high-precision sensor data to meet the needs of high-dynamic-range motion recognition.

This is a micro development board launched by Seeed Studio, based on the nRF52840 processor from Nordic Semiconductor. It uses the ARM Cortex-M4 core, offering powerful computational capabilities, and integrates Bluetooth 5.0 (Bluetooth Low Energy, BLE) wireless communication, meeting the needs of many low-power wireless devices. The development board is compact, with dimensions of 22mm x 17.5mm, making it suitable for embedded applications with limited space. Its core processor provides efficient computational performance, suitable for running computation-intensive tasks like Tiny ML models, while also offering low-power operation. In this project, the hardware mainly consists of the LSM6DS3 and XIAO-nRF52840 Sense.

The software part mainly includes the Arduino program for data uploading and the training of the Tiny ML model. Arduino has been widely used in the microcontroller field, so it will not be discussed in detail here. Edge Impulse is an edge AI platform specifically designed for embedded and IoT devices, aiming to deploy machine learning models on resource-constrained devices. The platform provides a comprehensive suite of tools and services for data collection, model training, and deployment, allowing developers to manage and implement machine learning tasks through a graphical interface, simplifying the machine learning workflow.

\section{Method}

In this project, the data collected primarily comes from the IMU sensor, specifically including acceleration data along the three axes: X, Y, and Z. This data reflects the acceleration changes of an object in three-dimensional space and is used to recognize and classify different activity patterns. We collected data for four different activity types: Walking, Stationary, Going Upstairs, and Going Downstairs.

For each activity, we collected 10 sets of data, with each set lasting 10 seconds. This results in a total of 40 sets of data, amounting to 400 seconds of data collection.

To train and validate the machine learning model, we divided the data for each activity into training and testing sets. From the 10 sets of data for each activity, 8 sets were randomly selected as the training set to train the neural network model, while the remaining 2 sets were used as the testing set to evaluate the model's performance and accuracy. In this way, we obtained a total of 32 training sets and 8 testing sets, with the training set comprising 80\% and the testing set comprising 20\%. This data division ratio follows the common standards used in machine learning.

Data pre-processing is a crucial step in machine learning as it helps convert raw data into a form that is more suitable for model training. Particularly when the raw data collected from the IMU accelerometer is time-series data, directly using this data as input for the model might result in poor model performance. Therefore, the primary task of data pre-processing is feature extraction, which involves extracting features from the time-series data that can effectively represent the movement patterns.

In this project, the sliding window technique was used to process the time-series data. The 10-second time series for each activity was divided into smaller time windows. Specifically, a 2-second window was used, with each window sliding every 80 milliseconds. This method allowed us to break down the continuous time series into multiple independent samples, so each original 10-second sample was split into 101 smaller samples.

Once the data samples were obtained through the sliding and overlapping windows, the next step was to perform spectral analysis. Spectral analysis is a technique used to extract important features from time-series data, helping to better understand and describe the characteristics of the signals. Several common feature extraction methods were applied, such as RMS (Root Mean Square) and FFT (Fast Fourier Transform). RMS reflects the average energy of the signal, helping to identify the signal's strength, while FFT analyzes the frequency components of the signal, revealing the frequency spectrum characteristics of different movement states. These features are highly useful in distinguishing different activity modes, as the frequency characteristics of walking and stationary states are noticeably different.

For classification, a simple Dense Neural Network (DNN) was used. This network consists of four layers of neurons and is designed to accurately classify four types of activities: Walking, Stationary, Going Upstairs, and Going Downstairs. The input layer consists of 39 neurons, each representing one of the 39 features extracted from the IMU sensor data. These features, including the mean and standard deviation of acceleration and angular velocity, were derived from the data preprocessing step and are aimed at capturing the dynamic characteristics of each activity.

The network structure includes two hidden layers: the first hidden layer has 20 neurons, and the second hidden layer has 10 neurons. These hidden layers serve to progressively abstract and transform the input features, allowing the neural network to capture complex nonlinear relationships in the data. Each neuron in these layers is processed through an activation function, commonly using ReLU (Rectified Linear Unit), which introduces nonlinearity, enabling the network to learn more complex patterns.

\section{Result}

\begin{figure*}[h]
	\centering
		\includegraphics[scale=1]{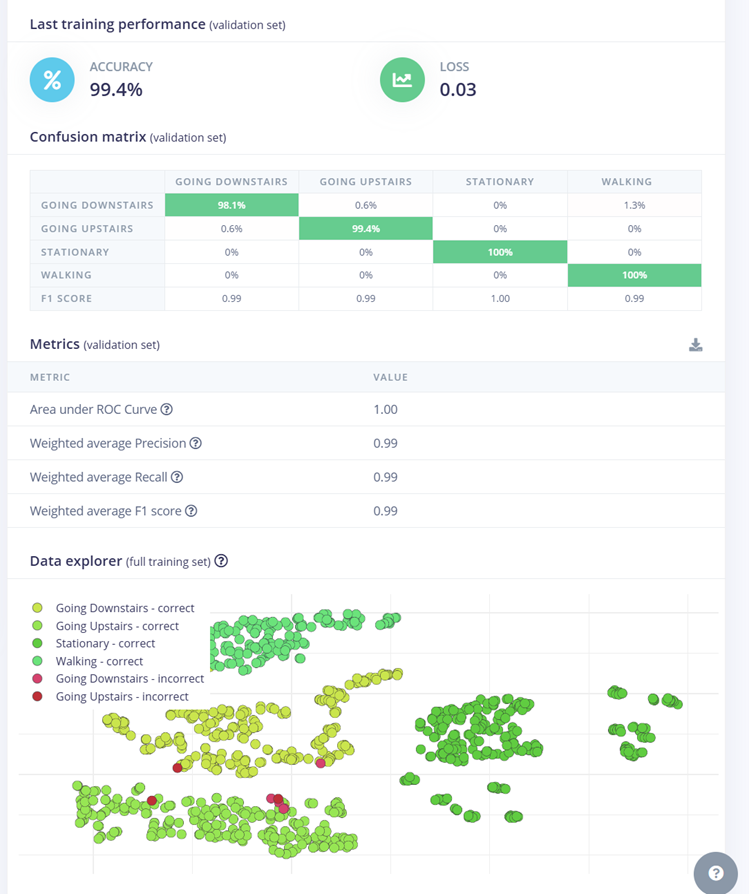}
	  \caption{Initial accuracy after training the model with the training set}
      \label{FIG:1}
\end{figure*}

\begin{figure*}[h]
	\centering
		\includegraphics[scale=1]{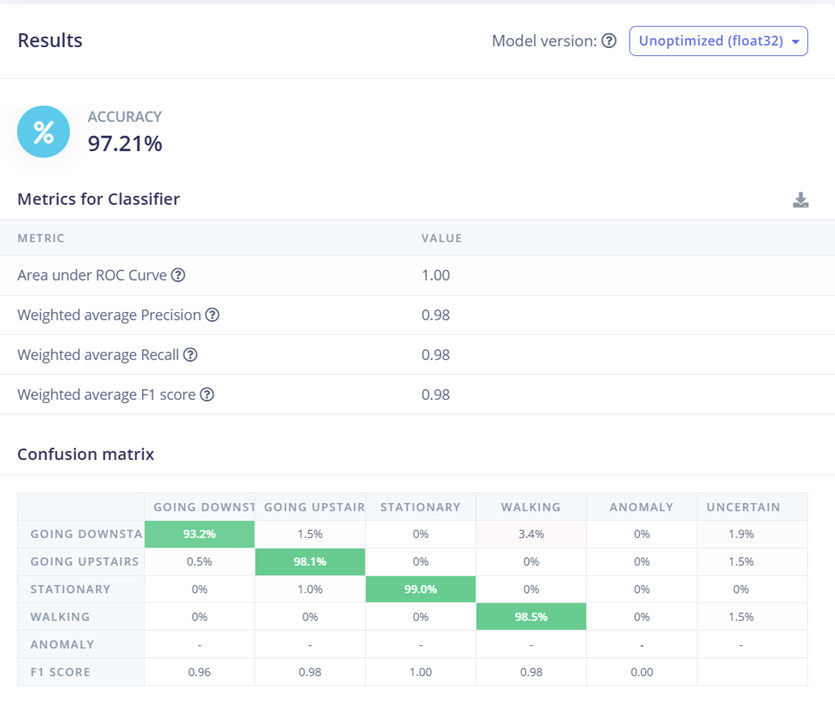}
	  \caption{The model accuracy measured using the test set}
      \label{FIG:2}
\end{figure*}

\begin{figure*}[h]
	\centering
		\includegraphics[scale=1]{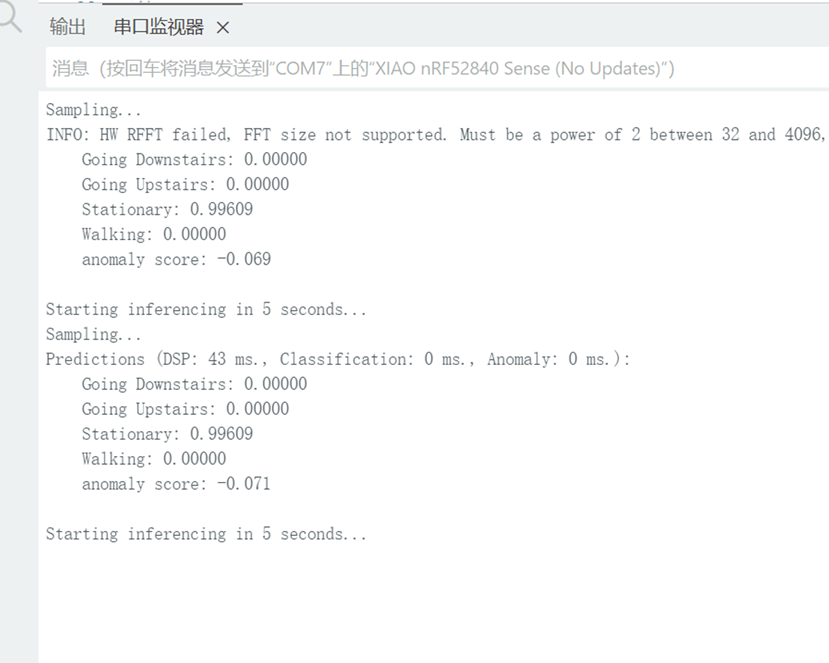}
	  \caption{Serial monitor results}
      \label{FIG:3}
\end{figure*}

Figure 1 shows the training performance of the model on the validation set. First, looking at the accuracy and loss value, the model achieves an accuracy of 99.4\%, and the loss value is 0.03, indicating that the model has trained very well and its predictions are very close to the actual values. The low loss value suggests that the model's error during training is small, and it has learned the important patterns in the data.
The confusion matrix in the figure displays the model’s prediction performance across various categories (Going Downstairs, Going Upstairs, Stationary, Walking). In the Going Downstairs category, the model correctly predicted 98.1\% of the samples, with incorrect predictions occurring as Walking and Going Upstairs. For the Going Upstairs category, the model’s correct prediction rate was 99.4\%, with errors primarily occurring in the Going Downstairs category. For the Stationary category, the model made perfect predictions, with 100\% of the samples correctly classified.

The F1 score, which considers both precision and recall, is close to 1 for all categories, with the Stationary category achieving a perfect score of 1.0, meaning the model showed a balanced performance in predicting each category.The Area Under the ROC Curve (AUC) is 1.00, indicating that the model can perfectly distinguish between the different categories.The data exploration graph at the bottom shows the distribution of the training set, where green points represent samples correctly classified by the model, while red and yellow points represent incorrectly classified samples. We can observe that most data points are accurately classified, with only a few errors occurring between the Going Downstairs and Going Upstairs categories.

Figure 2 shows the model's accuracy after testing with the test set. The model's accuracy is 97.21\%, which is slightly lower compared to the previous result. Although the accuracy is still high, the model's prediction performance has shown a slight decline in some categories, particularly in the Going Downstairs category, where the accuracy dropped to 93.2\%, while the Walking category accuracy is 98.5\%. Additionally, the predictions for the Anomaly and Uncertain categories in the second figure are weaker, with lower accuracy. This might be due to the limited data for these categories or their difficult-to-differentiate nature. However, an accuracy of 97.21\% is still considered very high performance.

Figure 3 shows the real-time inference results using the serial monitor after deploying the trained model onto the microcontroller. This is an example of data from a stationary state. For the four activity categories, the model's predicted results are as follows: the predicted probabilities for Going Downstairs and Going Upstairs are both 0.00000, indicating that the model did not recognize the current data as either of these activities. The predicted probability for Stationary is 0.99609, meaning the model is highly confident that the current data belongs to the stationary state, with nearly 100\% accuracy. For Walking, the model's predicted probability is 0.00000, indicating it did not classify the data as walking.Additionally, the anomaly score is -0.069, indicating that the model considers the current data to be normal, as the score is low.

\section{Conclusion}

This project successfully implemented a gait recognition system based on Tiny Machine Learning and IMU sensors. By using the XIAO-nRF52840 Sense microcontroller and LSM6DS3 IMU sensor, combined with the Edge Impulse platform for data collection, processing, and model training, the system is able to classify four common activities (Walking, Stationary, Going Upstairs, Going Downstairs) in real time. The data pre-processing step involved extracting relevant features using the sliding window technique, and the model was trained using a Deep Neural Network (DNN) classifier, achieving an accuracy of over 95\% on the test set. Additionally, the system includes anomaly detection functionality, which further enhances the robustness of the gait recognition system. The application of Tiny ML ensures that the system operates with low power, making it suitable for long-term, energy-efficient real-time monitoring.

Although the system has achieved good initial results in gait recognition tasks, there is still room for improvement. Future work will focus on the following aspects: First, expanding the range of activity recognition by adding more motion patterns and anomaly detection. Second, optimizing the existing model's accuracy and response time to further improve the system's performance in complex environments. Finally, integrating an energy-harvesting module to move the system towards full self-sufficiency, enhancing its feasibility and applicability in real-world use cases. With these improvements, it is expected that the system will play a larger role in areas such as health monitoring and smart homes.

% Use the style of numbering in square brackets.
% If nothing is used, default style will be taken.
%\begin{enumerate}[a)]
%\item 
%\item 
%\item 
%\end{enumerate}  

%\begin{itemize}
%\item 
%\item 
%\item 
%\end{itemize}  

%\begin{description}
%\item[]
%\item[] 
%\item[] 
%\end{description}  

%Figure
% \begin{figure}[h]
% 	\centering
% 		\includegraphics[scale=1]{elsevier-cas-double_column/wang_ruihua/f1.png}
% 	  \caption{Cardio axis}\label{fig:2}
% \end{figure}

% \begin{table}[h]
% \caption{Comparison of mechanical properties}\label{tbl1}
% \begin{tabular*}{\tblwidth}{@{}LL@{}}
% \toprule
%  Against Copper Film Test & Against Human Skin Test \\ % Table header row
% \midrule
%  Wet (10$\Omega$) & Wet (550$\Omega$) \\
%  Microneedle (1$\Omega$) & Microneedle (600$\Omega$) \\
%  Dry (0.1$\Omega$) & Dry (700$\Omega$) \\
% \bottomrule
% \end{tabular*}
% \end{table}

% Uncomment and use as the case may be
%\begin{theorem} 
%\end{theorem}

% Uncomment and use as the case may be
%\begin{lemma} 
%\end{lemma}



\bibliographystyle{cas-model2-names}

\bibliography{cas-refs}

% % Here goes the biography details.

% % Here goes the biography details.

\end{document}